\documentclass[letterpaper, 10 pt, conference]{ieeeconf}  

\IEEEoverridecommandlockouts                              
\title{\LARGE \bf
Trading the Twitter Sentiment with Reinforcement Learning
}
\usepackage{hyperref}
\usepackage{graphicx}

\author{Catherine Xiao \\  \href{mailto:Catherine.xiao1@gmail.com}{catherine.xiao1@gmail.com} 
   \and Wanfeng Chen \\ \href{mailto:wanfengc@gmail.com}{wanfengc@gmail.com} }

\begin{document}
\maketitle

\begin{abstract}

This paper is to explore the possibility to use alternative data and artificial intelligence techniques to trade stocks. The efficacy of the daily Twitter sentiment on predicting the stock return is examined using machine learning methods. Reinforcement learning(Q-learning) is applied to generate the optimal trading policy based on the sentiment signal. The predicting power of the sentiment signal is more significant if the stock price is driven by the expectation on the company growth and when the company has a major event that draws the public attention. The optimal trading strategy based on reinforcement learning outperforms the trading strategy based on the machine learning prediction.

\end{abstract}

\section{INTRODUCTION}

In a world where traditional financial information is ubiquitous and the financial models are
largely homogeneous, finding hidden information that has not been priced in from alternative
data is critical. The recent development in Natural Language Processing provides such
opportunities to look into text data in addition to numerical data. When the market sets the stock price, it is not uncommon that the expectation of the company growth outweighs the company fundamentals. Twitter, a online news and social network where the users post and interact with messages to express views  about certain topics, contains valuable information on the public mood and sentiment. A collection of research \cite{bollen_twitter_2011} \cite{si_exploiting_2013} have shown that there is a positive correlation between the "public mood" and the "market mood". Other research\cite{bollen_twitter_2011-1} also shows that significant correlation exists between the Twitter sentiment and the abnormal return during the peaks of the Twitter volume during a major event.

Once a signal that has predicting power on the stock market return is constructed, a trading strategy to express the view of the signal is needed. Traditionally, the quantitative finance industry relies on backtest, a process where the trading strategies are tuned during the simulations or optimizations. Reinforcement learning provides a way to find the optimal policy by maximizing the expected future utility. There are recent attempts from the Artificial Intelligence community to apply reinforcement learning to asset allocation \cite{moody_learning_2001}, algorithmic trading\cite{cumming_investigation_2015}\cite{varon_stock_2016}, and portfolio management\cite{jiang_deep_2017}.

The contribution of this paper is two-fold: First, the predicting power of Twitter sentiment is evaluated. Our results show sentiment is more suitable to construct alpha signals rather than total return signals and shows predicting power especially when the Twitter volume is high. Second, we proposed a trading strategy based on reinforcement learning (Q-learning) that takes the sentiment features as part of its states.

The paper is constructed as follows: In the second section, scraping Tweets from Twitter website and preprocessing the data are described in details. In the third section, assigning sentiment scores to the text data is discussed. In the fourth section, feature engineering and prediction based on the sentiment score is discussed. In the fifth section, how the reinforcement learning is applied to generate the optimal trading strategy is described.

\section{Twitter Data Scraping and Preprocessing}

There are two options of getting the Tweets. First, Twitter provides an API to download the Tweets. However, rate limit and history limit make it not an option for this paper. Second, scrapping Tweets directly from Twitter website. Using the second option, the daily Tweets for stocks of interest from 2015 January to 2017 June were downloaded. 

The predicting power of Twitter sentiment varies from stock to stock. For stocks that are mostly driven by the company fundamentals and hold by the institutional investors, the predicting power of the Twitter sentiment is limited. For stocks that are priced by the public expectation on the company's future growth, Twitter sentiment describes the confidence and expectation level of the investors. 

For this reason, two companies from the same industry, Tesla and Ford are investigated on how Twitter sentiment could impact the stock price. Tesla is an electronic car company 
that shows consecutive negative operating cash flow and net income but carries very high expectation from the public. Ford, is a traditional auto maker whose stock prices has been stabilized to represent the company fundamentals. 

To investigate how different key words impact the predicting power of the sentiment score, two Tweet sets, a ticker set and a product set, are prepared for each stock. The first set of Tweets are searched strictly according to the stock ticker. The second set of Tweets are searched according to the company's products and news. The keywords for the second dataset are defined according to the top twenty related keywords of the stock ticker according to Google Trend, a web facility shows how often a certain word is searched relative to Google's total search volume. For example, "Elon Musk" is among the set of keywords that retrieve the second tweets set for Tesla.

Tweets contain irregular symbols, url and emoji etc which has to be preprocessed so
that the NLP algorithm can extract the relevant information efficiently. Examples of preprocessing are described as below:
\begin{itemize}

\item Filter out tweets that contains “http” or ‘.com’
Motivation: They're usually ads
e.g. ‘\#Fhotoroom \#iPhone
https://www.fhotoroom.com/fhotos/’

\item Remove \#hashtag, @user, tabs and extra spaces
\item Filter out tweets that contain consecutive two question marks (‘?’).
The reason is the coding of these tweets is usually not recognizable.

\item Filtering out none-English tweets using Google’s Langdetect package.
\end{itemize}
\section{Sentiment Score}
\begin{figure}[htbp]
\centering
\includegraphics[width=0.4\textwidth]{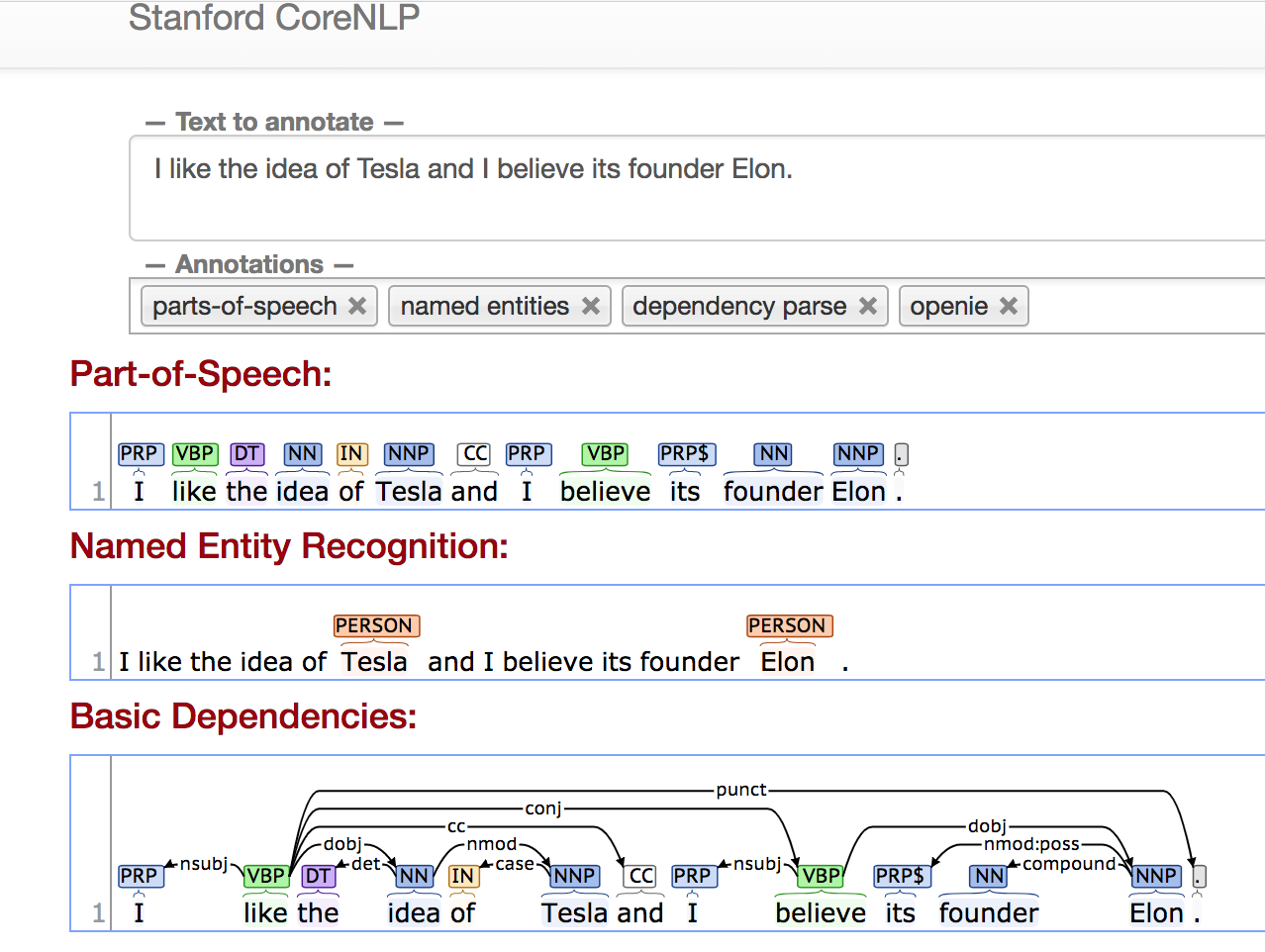}

\caption{Demo of the functionalities provided by Stanford CoreNLP}
\label{fig:nlp}
\end{figure}

To translate each tweet into a sentiment score, the Stanford coreNLP software was used. Stanford CoreNLP is designed to make linguistic analysis accessible to the general public. It provides named Entity Recognition, co-reference and basic dependencies and many other text understanding applications.
An example that illustrate the basic functionality of Stanford coreNLP is shown in Figure.\ref{fig:nlp}

coreNLP can compute a sentiment score for each sentence with value ranging from 0 to 4
, where 0 stands for negative, and 4 stands for very positive.
For tweets with multiple sentences, the average of the sentiment scores of
all sentences is used as the sentiment score of the Tweets.

The number of Tweets varies everyday from a couple of hundreds to over ten thousands, depends on if the company has a major event that attracts the public attention. The sentiment scores are normalized between 0 to 1, and features based on the sentiment score is constructed and normalized. 

\begin{figure}[htbp]
\centering
\includegraphics[width=0.5\textwidth]{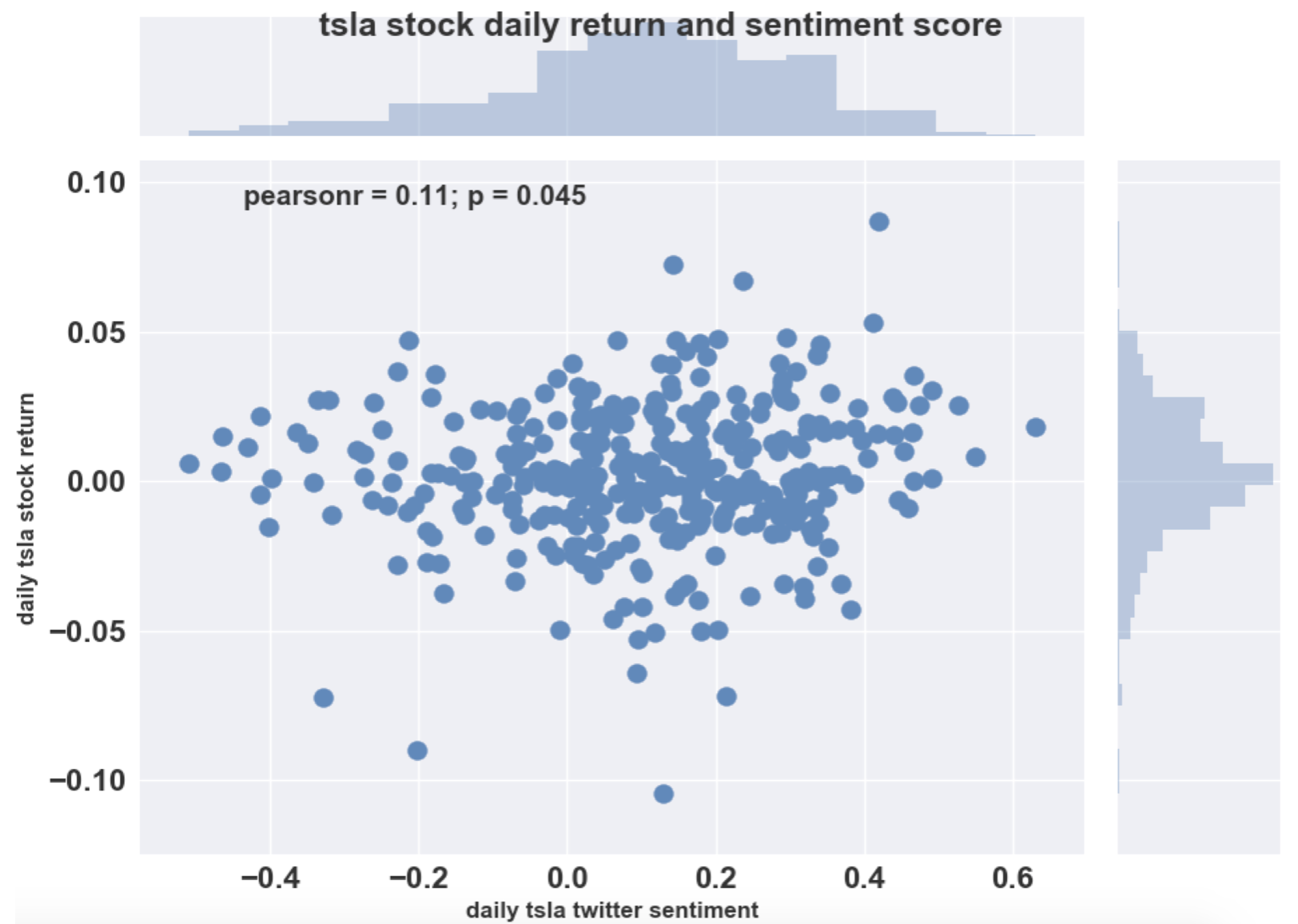}

\caption{Histogram}
\label{fig:scat}
\end{figure}

Figure \ref{fig:scat} shows the relationship between Tesla stock return and stock sentiment score. According the distribution of the sentiment score, the sentiment on Tesla is slightly skewed towards positive during the testing period. The price has been increased significantly during the testing period, which reflected the positive sentiment. The predicting power of sentiment score is more significant when the sentiment is more extreme and less so when the sentiment is neutral.

\section{The Sentiment machine learning model}
\subsection{Feature Engineering}
Feature engineering is the process to extract meaningful information from the raw data in order to improve the performance of machine learning mode. Domain knowledge and intuition are often applied to keep the number of the features reasonable relative to the training data size.
Two categories of features are defines: technical features and sentiment features. The technical features include previous day's return and volume, price momentum and volatility. The sentiment features include number of Tweets, daily average sentiment score, cross-section sentiment volatility, sentiment momentum and reversal.

\subsection{Machine Learning Prediction Model}

The logistic regression with L1 regularization and RBF-kernel SVM are applied to predict a binary outcome, i.e. whether the stock return will be positive or negative in the next day. Both technical and sentiment-based features carry important information about the stock price and therefore are provided as the model inputs. Half of the dataset is used for training and the rest is used for testing.

The 3 fold cross validation is applied to learn the model hyper-parameters. Specifically, the
hyper-parameters C of both models and γ of RBF-kernel SVM are learned such that the dev set accuracy is maximized.
The hyper-parameter C  in logistic regression determines the degree of regularization. Smaller C means more regularization, i.e. high bias and low variance. RBF-kernel SVM has two hyper-parameters, C and γ. C controls the width of soft margin, smaller C allows placing more samples on the wrong side of the margin. γ is a parameter in RBF kernel. A larger γ means a Gaussian with smaller variance and thus less influence of support vectors. Typically, small C and large γ lead to high bias and low variance.

To evaluate if the sentiment feature improves the prediction accuracy, a baseline model is defined. The baseline applies linear logistic regression to a set of stock technical signals to predict the following day’s stock return sign (+/‐). No sentiment features are provided to the baseline model.

\begin{figure}[htbp]
\centering
\includegraphics[width=0.5\textwidth]{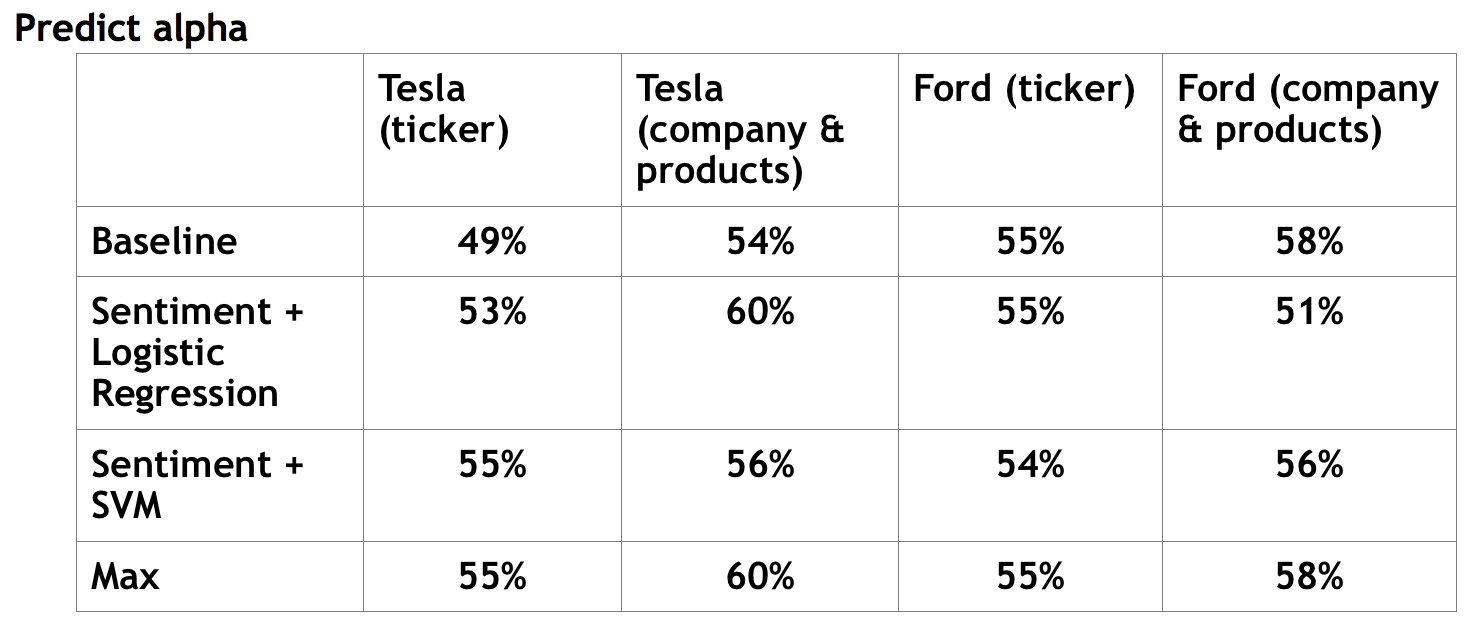}

\caption{The chart displays the accuracy on predicting the "alpha", which defines as the return of the stock minus the return of its sector ETF. }
\label{fig:rr}
\end{figure}

\subsection{Predicting using ticker dataset and product dataset}
The predicting power for the ticker dataset and product dataset are compared. The ticker dataset contains tweets that searched strictly according to the stock ticker. The product dataset is searched using keywords that related to the company's product and other related topic(see session II for more details).
The former dataset represents the investors' sentiment, while the latter dataset represents customers’ sentiment.

In the Tesla case, using product tweets consistently outperforms using the ticker tweets(accuracy 0.6 vs 0.5), it is less so in the Ford case(0.58 vs 0.55). The result is displayed in Figure \ref{fig:rr} First, this is because Tesla's stock price is driven more by the sentiment on its product instead of the stock itself. For Ford, not many people actually express their opinion about Ford's product via Twitter. Secondly, Tesla has many more product tweets than ticker tweets, but Ford is opposite. 

\subsection{Predicting using logistic regression and SVM}
In most cases, SVM performs only slightly better than logistic regression in validation set, although much better in testing set. This may be because the dataset is not large enough to prevent SVM overfitting. The comparision between the logistic regression and the SVM is displayed in Figure \ref{fig:rr}

\subsection{Predicting Total Return vs Alpha}
It is important to identify which is a better target for the prediction. Two targets, predicting "alpha or predicting "total return" are compared. "Alpha" defines as the excess stock return over its sector ETF. "Total return" is the absolution stock return.
Predicting "alpha" achieves better performance than predicting total return. This is because the sentiment is more related to stock’s idiosyncratic. Good sentiments towards a specific company or its stock won’t override the overall stock market or sector’s impact on the stock return. 

\subsection{Tesla vs Ford}
The prediction accuracy on Tesla is higher than Ford according to Figure\ref{fig:rr}. The reason is because Tesla's stock price largely reflects the sentiment and confidence level of the public. The company has consecutive negative cash flow and net income, making prediction based on its fundamental information unrealistic. On the other hand, the stock price of Ford, which is a traditional automaker, is not that related to the public sentiment.

\subsection{Feature selection and overfitting}
\begin{figure}[htbp]
\centering
\includegraphics[width=0.5\textwidth]{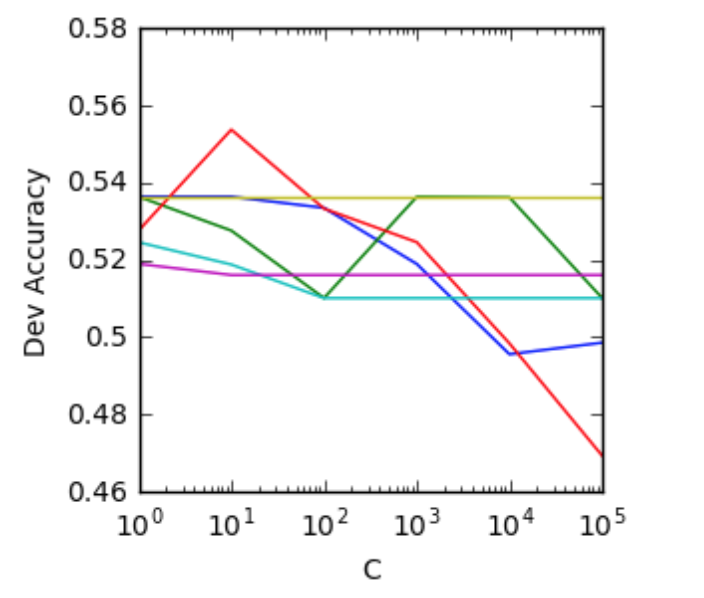}

\caption{The chart shows an example of overfitting in the SVM model. The overfitting is caused by adding too many features to the model inputs but not providing enough data for the model to generalize. Different lines shows the SVM performance under different $\gamma$ parameter. None of the parameter achieves better accuracy than a restricted set of features.}
\label{fig:svm2}
\end{figure}

To improve the model accuracy, more features were constructed. However, more features do not result in better accuracy. For example, in Figure \ref{fig:svm2}, adding more features improve the training accuracy but deteriorates out-of-sample accuracy due to overfitting. 

The recursive feature elimination and cross validation (RFECV) for feature selection is experimented during the feature selection phase. However, only similar or even slightly worse performance was achieved by RFECV than selecting features according to domain knowledge and intuition. This is because recursive feature elimination is a greedy algorithm and thus doesn’t guarantee optimal solution.

\section{Q-learning}

Q-learning is a model-free reinforcement learning technique. Specifically, Q-learning can be used to find an optimal policy given a Markov Decision Process(MDP). Instead of learning the transition probability, Q-learning directly learns the expected utility of taking an action from a certain state. By maximizing the expected utility of the certain state, the optimal policy is found.

Traditionally, quants propose trading strategies according to backtest, where the optimal parameters are tuned by maximizing the objective function based on historical data. However, this common practice adopted by the investment industry has drawbacks. First, it over-fits the historical data and doesn't generalize to out of sample data. In addition, the model need to be recalibrated periodically due to the economic regime change. A strategy significantly outperforms in a high volatility environment might suffer significantly in a low volatility environment. 

The Q-learning, in the opposite, learns from the feedback from the market, and generates the optimal trading strategy according to past experience, and automatically adapts to the new market regime. 

In this paper, the Q-learning algorithm is applied to generate the optimal trading strategy. The market is modeled as a Markov Decision Process where the outcomes are random and not under the control of the decision maker. The states contain information of three categories: technical indicators, sentiment features and portfolio information. The actions contains buy, sell and hold. The reward is the next day market return. The limit of leverage and the loss-cutting threshold are implemented in the relation ship of successor state and action. For example, if the leverage constrain has been met, the actions that valid for this state are only "hold" or "sell". If the loss cutting threshold has been triggered, say the portfolio lost half of the capital and this is the maximum tolerance of loss, only the action that exit current position is valid. 
\subsection{Learning}
Formally, the learning process defines as below. In Q-learning the optimal expected utility of a (state, action) pair $\hat{Q}_{opt}(s, a)$ is updated with the rewards $r$ and the expected utility of the subsequent state $\hat{V}_{opt}(s')$ after taking the action $a$. 
\begin{equation}\label{qopt}
\hat{Q}_{opt}(s, a)\leftarrow (1-\eta) \hat{Q}_{opt}(s, a)+ \eta (r + \gamma \hat{V}_{opt}(s')
\end{equation}
\begin{equation}\label{vopt}
V_{opt}(s') = \max_{a' \in Actions(s')}\hat{Q}_{opt}(s',a')
\end{equation}

The optimal policy is proposed by Q-learning as 
\begin{equation}\label{optimal a}
\pi_{opt}(s)=\arg\max\limits_{a\in act(s)} Q_{opt}(s,a)
\end{equation}

\subsection{Function Approximation}

Function approximation refers to the method to generalize unseen states by applying machine learning methods.  The Q-table stores the expected utility for each (state,action) pair that has been explored. When predicting the expected utility for a certain (state, action) pair, we will look up the Q-table. When the MDP has many states and actions, it is very likely that a (state, action) pair has not been explored yet so the estimate is not accurate. It is too slow to look up a gigantic table and most likely there is not enough training data to learn each of the state individually. Function approximation uses features to capture the characteristics of the states and applies stochastic gradient descent to update the weights on each feature.  More specifically, below equation is applied to generalize the unseen state in this paper.
Define features $\phi(s,a)$ and weights $w$, then
\begin{equation}\label{FA}
\hat{Q}_{opt}(s, a; w) = w \cdot \phi(s, a)
\end{equation}
For each $(s,a,r,s')$, apply stochastic gradient descent to update the weights.

\begin{equation}\label{FA}
w \leftarrow w - \eta [\hat {Q}_{opt}(s, a; w)- (r + \gamma \hat{V}_{opt}(s'))]\phi(s, a)
\end{equation}
where $\eta$ is the learning rate, $r$ is the reward and $\gamma$ is the discount factor.

\subsection{Exploration and Exploitation}
It is necessary to balance the exploration and exploitation. One might suggest naively to take action only according to the optimal policy estimated by maximizing $\hat{Q}_{opt}(s,a)$. However, this greedy strategy is equivalent to stay in the comfortable zone all the time in life, without gaining new experience and unable to give reasonable prediction when encounters unseen situations. Another extreme is to always explore by choosing an action randomly. Without applying the hard lesson learned and obtaining the rewards, the algorithm can lead to unsatisfiable utility at the end. Therefore, in this paper the Epsilon-greedy strategy is applied for exploration. For a certain probability, the algorithm acts randomly(exploration), for the rest the algorithm acts optimally(exploitation). 
\subsection{Result and Discussion}
\begin{figure}[htbp]
\centering
\includegraphics[width=0.5\textwidth]{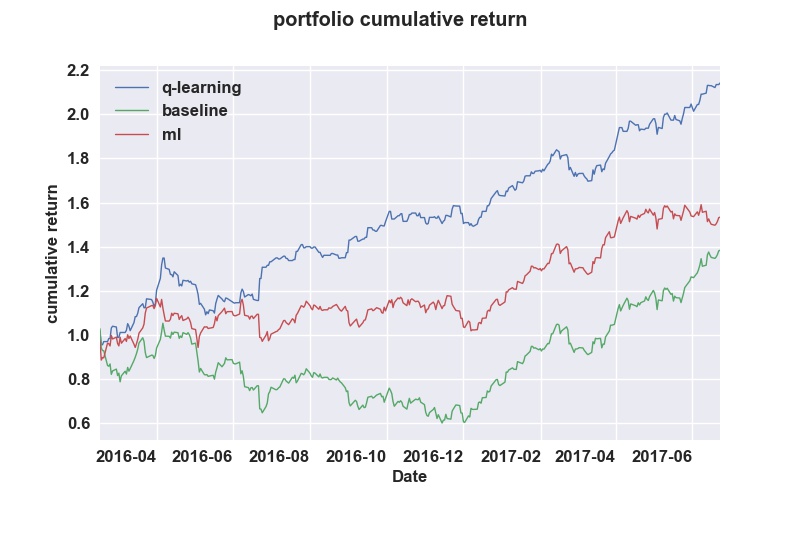}

\caption{The chart shows the trading strategy derived from Q-learning(in blue) outperform the backtest result using machine learning features(in red). Both of Q-learning strategy and machine learning strategy outperform the baseline(in green). }
\label{fig:rtn}
\end{figure}

Figure \ref{fig:rtn} shows the cumulative return over 1 year period. The strategy trades daily.
The Q-learning states include portfolio position, sentiment features and technical indicators such as price momentum. The machine learning strategy predicts the binary movement (+ or -) of next trading day price based on sentiment features and technical indicators. The backtest rule based on the machine learning prediction is to long the stock if the prediction is +, short the stock if -. The baseline is the same with machine learning except only the technical indicator was used as the feature. The oracle model of this project is a trader who has insider information about the stock and be able to bet and act correctly on every single day of the testing period. The oracle model is able to achieve 6 times of the initial capital at the end of testing period.

There are observations that worth a discussion. At the beginning of the testing period, the Q-learning has not learnt how to estimate the expected utility of a certain action yet. The performance of the initial period is more unstable than later. Q-learning does better when the state is more common because it accumulates more experience about the situation but might not take the best action when a outlier state is presented. The performance of the q-learning varies during different batch due to the random nature of exploitation and exploration. In general Q-learning is able to deliver better performance than using the binary prediction from the machine learning models. Both of the Q-learning and machine learning model outperform the baseline model.

\section{Future work}
There are many areas that can be improved given more resource and data. Below is a list of the improvement that could make this idea more robust.
\begin{itemize}
\item Use intraday data to test the sentiment signal and Q-learning. By training with more data and trading more promptly, we expect both sentiment machine learning model and Q-learning to do better.
\item With more data, more features can be considered and incorporated into the model.
\item Apply different function approximators, for example, neural net, to better generalize the states and provide more stable behavior

\item Add another class to the existing binary classifier – insignificant price change. The is motivated by preventing the classifier to fit to the noise inherent in stock market price movement, and lumps small, statistically insignificant upward or downward movements indiscriminately with large ones.
\item Add crude oil future price as a feature to predict Tesla stock alpha return sign

\item Extend the sentiment analysis to other stocks and Cryptocurrency, which is an asset class that driven even more by the public sentiment.

\end{itemize}
\section{CONCLUSIONS}

The paper explores the possibility to predict stock price using text data and reinforcement learning technique. Predicting stock price direction using Twitter sentiment is challenging but promising. Which stock and what to predict is more important than how to predict. For example, Tesla, a company driven by the expectation of the company's growth is a better target than Ford, a traditional auto maker. Reinforcement learning is applied to find the optimal trading policy by learning the feedbacks from the market. The Q-learning is able to adapt automatically if the market regime shifts and avoid backtesting, a process applied by investment industry that often overfit the historical data. Both of the machine learning model and the Q-learning model outperforms the baseline model, which is a logistic regression without sentiment features.
\addtolength{\textheight}{-12cm}   


%
%



\section*{ACKNOWLEDGMENT}

We would like to thank Anna Wang, who is the project mentor, gives very practical suggestions and guidance. We would like to thank Standford University for the very challenging and exciting CS221 course materials and Prof. Percy Liang who did such a great job getting us interested in sentiment analysis and reinforcement learning

\bibliographystyle{plain}
\bibliography{My_Library.bib}

\begin{thebibliography}{1}

\bibitem{bollen_twitter_2011}
Johan Bollen and Huina Mao.
\newblock Twitter mood as a stock market predictor.
\newblock {\em Computer}, 44(10):91--94, 2011.

\bibitem{bollen_twitter_2011-1}
Johan Bollen, Huina Mao, and Xiao-Jun Zeng.
\newblock Twitter mood predicts the stock market.
\newblock {\em Journal of Computational Science}, 2(1):1--8, March 2011.
\newblock arXiv: 1010.3003.

\bibitem{cumming_investigation_2015}
James Cumming, Dalal Alrajeh, and Luke Dickens.
\newblock An {Investigation} into the {Use} of {Reinforcement} {Learning}
  {Techniques} within the {Algorithmic} {Trading} {Domain}.
\newblock 2015.

\bibitem{jiang_deep_2017}
Zhengyao Jiang, Dixing Xu, and Jinjun Liang.
\newblock A {Deep} {Reinforcement} {Learning} {Framework} for the {Financial}
  {Portfolio} {Management} {Problem}.
\newblock {\em arXiv preprint arXiv:1706.10059}, 2017.

\bibitem{moody_learning_2001}
J.~Moody and M.~Saffell.
\newblock Learning to trade via direct reinforcement.
\newblock {\em IEEE Transactions on Neural Networks}, 12(4):875--889, July
  2001.

\bibitem{si_exploiting_2013}
Jianfeng Si, Arjun Mukherjee, Bing Liu, Qing Li, Huayi Li, and Xiaotie Deng.
\newblock Exploiting {Topic} based {Twitter} {Sentiment} for {Stock}
  {Prediction}.
\newblock {\em ACL (2)}, 2013:24--29, 2013.

\bibitem{varon_stock_2016}
Jonah Varon and Anthony Soroka.
\newblock Stock {Trading} with {Reinforcement} {Learning}.
\newblock 2016.

\end{thebibliography}

\end{document}